\documentclass[letterpaper]{article} 
\usepackage{aaai25}  
\usepackage{times}  
\usepackage{helvet}  
\usepackage{courier}  
\usepackage[hyphens]{url}  
\usepackage{graphicx} 
\usepackage{natbib}  
\usepackage{caption} 
\usepackage{algorithm}
\usepackage{algorithmic}

\usepackage{newfloat}
\usepackage{listings}
\DeclareCaptionStyle{ruled}{labelfont=normalfont,labelsep=colon,strut=off} 
\floatstyle{ruled}
\newfloat{listing}{tb}{lst}{}
\floatname{listing}{Listing}
\usepackage{booktabs}
\usepackage{color}
\usepackage{adjustbox}
\usepackage{array}
\usepackage{xcolor,colortbl}
\usepackage{appendix}
\usepackage{algorithm,algorithmic}
\usepackage{subfig}

\usepackage{multirow}
\usepackage{amsmath}
\usepackage{amssymb}

\newcommand{\Fmat}[0]{{{\textbf F}}}

\newcommand{\Hmat}[0]{{{\textbf H}}}

\newcommand{\Wmat}[0]{{{\textbf W}}}
\newcommand{\Xmat}{{\textbf X}}
\newcommand{\Ymat}[0]{{{\textbf Y}}}
\newcommand{\Zmat}{{\textbf Z}}

\newcommand{\xv}{\boldsymbol{x}}
\newcommand{\yv}{\boldsymbol{y}}

\newcommand{\gammav}{{\boldsymbol{\gamma}}}

\usepackage{xspace}

\title{Prior-guided Hierarchical Harmonization Network for \\ Efficient Image Dehazing}
\author{
Xiongfei Su\textsuperscript{\rm 1,2}, Siyuan Li\textsuperscript{\rm 1,2}, Yuning Cui\textsuperscript{\rm 3}, Miao Cao\textsuperscript{\rm 1,2}, Yulun Zhang\textsuperscript{\rm 4}, Zheng Chen\textsuperscript{\rm 4}, \\
Zongliang Wu\textsuperscript{\rm 1,2}, Zedong Wang\textsuperscript{\rm 2}, Yuanlong Zhang\textsuperscript{\rm 5}, Xin Yuan\textsuperscript{\rm 2}\thanks{Corresponding author.}
}
\affiliations{
    \textsuperscript{\rm 1}Zhejiang University, Hangzhou, China, \\ 
    \textsuperscript{\rm 2}Westlake University, Hangzhou, China \\
    \textsuperscript{\rm 3}Technical University of Munich, Munich, Germany \\
    \textsuperscript{\rm 4}Shanghai Jiao Tong University, Shanghai, China \\
\textsuperscript{\rm 5}Tsinghua University, Beijing, China \\

}

\usepackage{bibentry}

\begin{document}

\maketitle

\begin{abstract}
  Image dehazing is a crucial task that involves the enhancement of degraded images to recover their sharpness and textures. While vision Transformers have exhibited impressive results in diverse dehazing tasks, their quadratic complexity and lack of dehazing priors pose significant drawbacks for real-world applications.
  In this paper, guided by triple priors, Bright Channel Prior (BCP), Dark Channel Prior (DCP), and Histogram Equalization (HE), we propose a \textit{P}rior-\textit{g}uided Hierarchical \textit{H}armonization Network (PGH$^2$Net) for image dehazing.
  PGH$^2$Net is built upon the UNet-like architecture with an efficient encoder and decoder, consisting of two module types: (1) Prior aggregation module that injects B/DCP and selects diverse contexts with gating attention. (2) Feature harmonization modules that subtract low-frequency components from spatial and channel aspects and learn more informative feature distributions to equalize the feature maps.
  Inspired by observing the lower sparsity of B/DCP and the histogram equalization, we harmonize the deep features using a histogram equation-guided module and further leverage B/DCP to guide spatial attention through a sandwich module as the bottleneck.
  Comprehensive experiments demonstrate that our model efficiently attains the highest level of performance. 
https://github.com/Nicholassu/PGHHNet.
\end{abstract}

%
\section{Introduction}
\label{sec:intro}

Image dehazing aims to recover clear images from hazy ones~\cite{zheng2023curricular,cui2025eenet}. It is crucial in fields like surveillance, autonomous driving, and remote sensing. Estimating clean backgrounds, textures, and colors from a single hazy image is complex and ill-posed. Solutions fall into three categories: conventional, deep learning, and hybrid methods.
\begin{figure}[htbp!]
    \centering
    \includegraphics[width=\linewidth]{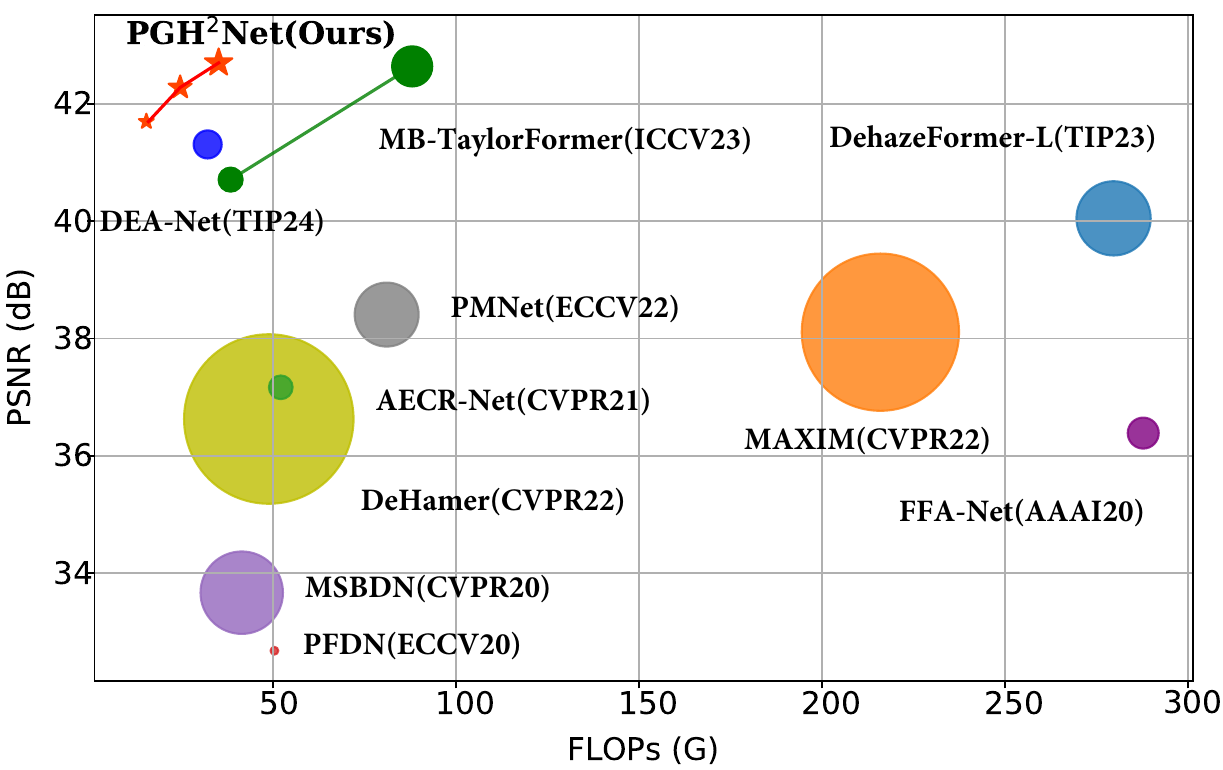}
    \caption{Reconstruction quality (PSNR) and computational complexity (FLOPs) on the SOTS-Indoor~\cite{RESIDE} dataset. The size of the dots indicates the model size.}
    \label{fig:bubble}
\end{figure}
\textbf{$(i)$ Conventional methods}~\cite{MPR,he2010single} rely on physical model assumptions and manual feature engineering, often failing in real-world situations due to the problem's ill-posed nature. These priors are only effective in specific scenarios, as handcrafted features are too simple for complex phenomena like haze and have difficulty selecting optimal transforms and tuning parameters.
\begin{figure*}[!t]
    \centering
    \includegraphics[width=0.95\linewidth]{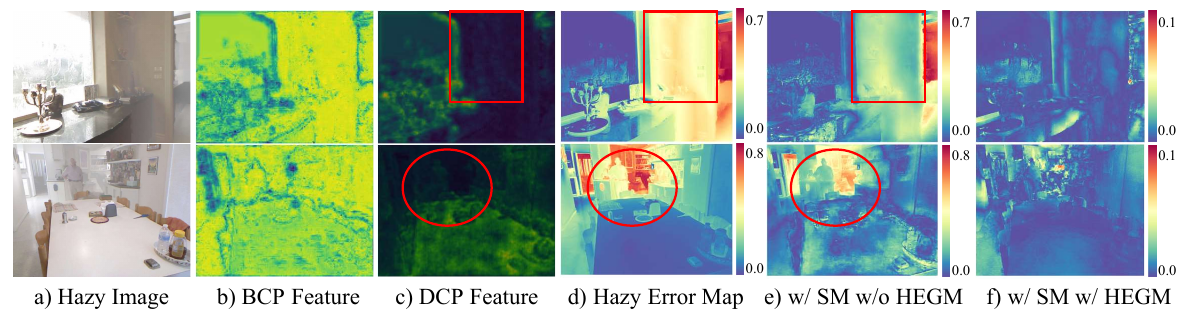}
    \caption{Visualization of the relationship between the spatial haze degradation(a)(d) and BCP(b) and DCP(c) in the deep feature domain. The error maps are differential values with reference, indicating haze distribution, shown in red boxes. 
    }
    \label{fig:motivation_SM}
\end{figure*}
\begin{figure*}[!htbp]
    \centering
    \includegraphics[width=0.95\linewidth]{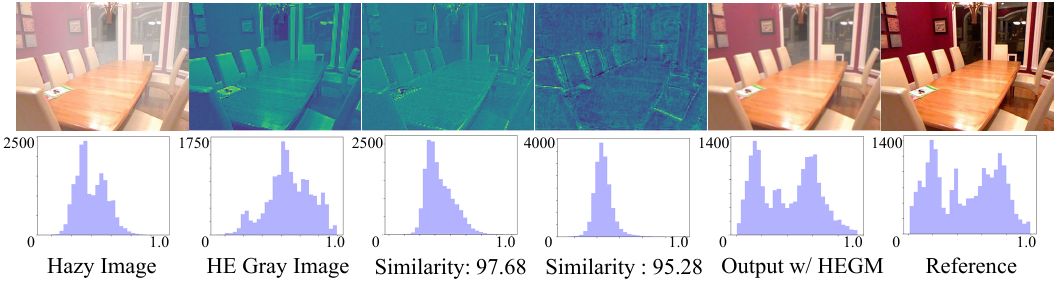}
    \caption{Visualization of the relationship between the value distribution and feature channels. The second row shows histograms assigned to the first row of each image/feature. Similarities are calculated by Cosine similarity. The horizontal/x-axis is the normalized value from 0 to 1, and vertical/y-axis is the number of the value distribution of images/feature. }
    \label{fig:motivation_HE}
\end{figure*}
\textbf{$(ii)$ Deep learning methods} use CNN-based approaches~\cite{bai2022self,cui2024revitalizing}, including encoder-decoder structures, dilated convolution, and attention mechanisms, achieving impressive restoration performance. These methods employ deep learning as a black box for image restoration but rely on 2D images, with computational time increasing rapidly with larger image sizes.
Recently, Transformer models have shown significant improvements in image restoration~\cite{Chen_2021_CVPR,song2023vision}, but these often increase model complexity, training costs, and convergence issues due to numerous parameters.
\textbf{$(iii)$ Hybrid methods}~\cite{zheng2023curricular, mo_dca-cyclegan_2022} reduce dependence on training data by combining inherent priors with the representation ability of deep neural networks.
\cite{cai_dark_2020, zheng2023curricular, dai2022deep} use physics priors in the feature space to enhance interpretability aligned with the hazing process. However, these priors are limited to shallow layers, which lose rich information in deep layers. Thus, universal priors and hierarchical mechanisms are necessary to advance hybrid methods.

In this paper, we empirically reveal the guidance mechanism of Bright Channel Prior (BCP) and Dark Channel Prior (DCP) in the hierarchical feature domain and the distribution matching mechanism of Histogram Equalization (HE).
Specifically, we first analyze the deep feature maps ($128\times64\times64$) of a basic UNet and split the feature maps into bright channel and dark channel. The error map in Fig.~\ref{fig:motivation_SM}(d) is the difference between input and reference images, indicating haze distribution. So red and blue are the thick and thin hazy regions, respectively. The distribution of DCP is basically consistent with the error map, which is sufficient to serve as coarse guidance for hazy regions. Fig.~\ref{fig:motivation_SM}(c) distinguishes the cleaner window and hazy wall (red box) in the top row,  as well as the cleaner table and hazy back room (red circle) in the bottom row. For BCP, it highlights the major high-frequency information, which is shown in Fig.~\ref{fig:motivation_SM}(b). However, limitations of B/DCP exist in Fig.~\ref{fig:motivation_SM}(e): The spatial guidance~\cite{DehazeMamba,yao2023bidirectional} with B/DCP can remove the major haze, but it is necessary to harmonize with the HE prior, verified in Fig.~\ref{fig:motivation_SM}(f) and Tab.~\ref{tab:ablation},\ref{tab:breakdown}.


To further explore the priors in hierarchical levels, we notice that distribution (such as histogram) with only one dimension vector is easy to transport in deep layers 
without the limitation of hierarchical sizes. In Fig.~\ref{fig:motivation_HE}, we first generate the histogram of the hazy image and processed image by HE. It is apparent that hazy images own more voxel values close to white, causing a peak in the histogram results. HE flattens the histogram with a remapping algorithm. Subsequently, we calculate the distribution similarity between each channel feature and HE. We show the distribution of two channels in Fig.~\ref{fig:motivation_HE} with similarities and find that similar distributions with HE indicate cleaner and sharper corresponding channel features.

Based on the above observation and analysis, we design a novel hierarchical pure convolutional architecture, {\em dubbed PGH$^2$Net}, for image dehazing tasks. From a fresh perspective, we solve the ill-posed problem by jointly introducing channel and distribution priors into deep layers of the network to guide the restoration from hierarchical levels:

\begin{figure*}[!ht]
    \centering
    \includegraphics[width=\linewidth]{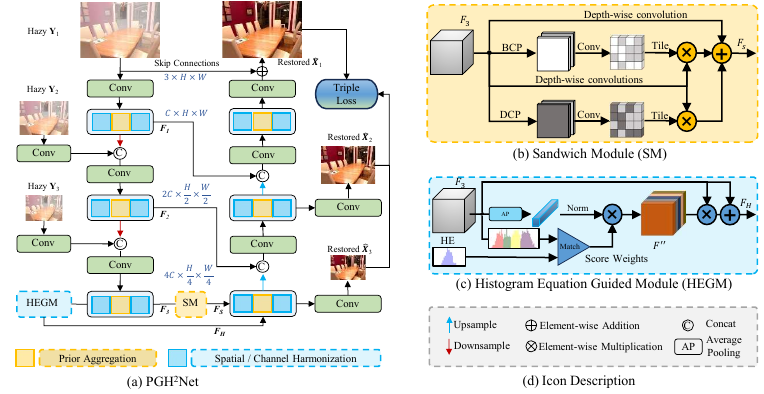}
    \caption{PGH$^2$Net architecture. (a) The encoders and decoders with a stack of Prior Aggregation and Spatial/Channel Harmonization modules learn hierarchical features with diverse distributions. Then, the bottleneck with the (b) Sandwich Module (SM) and (c) Histogram Equation Guided Module (HEGM) transports equalized deep features to the decoders.}

    \label{fig:pipeline}
\end{figure*}
\begin{itemize}
    \item We reveal the spatial guidance mechanism of B/DCP in the hierarchical feature domain and propose a design philosophy of \textbf{aggregating priors}. The Prior Aggregator injects B/DCP and selects diverse contexts via gating attention, while the Sandwich Module as bottleneck injects B/DCP with complementary spatial attention.
    \item We reveal the distribution guidance from HE prior as another principle, \textbf{harmonizing feature distributions}. Spatial and Channel Harmonization Modules enrich and equalize features by adaptively removing low-frequency components, while the HE Guidance Module as bottleneck provides channel-wise weighting harmonization.
    \item Extensive experiments demonstrate that the proposed PGH$^2$Net performs favorably against previous state-of-the-art algorithms. Meanwhile, PGH$^2$Net significantly reduces the computational complexity and achieves a sweet point in the performance-parameters trade-off.
\end{itemize}

\section{Related Work}
\label{sec:related}

\subsection{Image Dehazing}

Researchers have begun using deep neural networks for image dehazing~\cite{liu2019learning,song2023vision,bai2022self,zheng2023curricular,cui2023selective,cui2023focal}. CNN architectures significantly outperform traditional physical-based methods~\cite{DPDD,NAFNet}. The encoder-decoder paradigm, used for hierarchical representations~\cite{deeprft}, has been enhanced with modules like dynamic filters~\cite{IFAN}, dilated convolution~\cite{zou2021sdwnet}, shortcut connections~\cite{MIMO,cui2023image,cui2023dual}, and attention modules~\cite{qin2020ffa}. Vision Transformers have shown impressive results in image dehazing but suffer from quadratic complexity, leading researchers to restrict operation regions~\cite{Liang_2021_ICCV} or switch operation dimensions~\cite{zamir2022restormer}. Hybrid methods have also been introduced, combining statistics-based priors with deep learning, such as DCA-CycleGAN~\cite{mo_dca-cyclegan_2022}, which integrates dark channel prior.

\subsection{Triple Priors}
He \textit{et al.}~\cite{he2010single} introduced the dark channel prior, positing that its values in an unobstructed image tend to be close to zero. \cite{pan_deblurring_2018} verifies its effectiveness in deblurring tasks. However, this approach works well for most outdoor hazy images but struggles with hazy images featuring bright areas, especially in the sky. \cite{yan_image_2017} leverages B/DCP and incorporates a prior using both bright and dark information. \cite{cai_dark_2020} uses B/DCP in a multi-branch network layer to extract feature information, increasing computational complexity.

Histogram adjustment is another widely used prior, helpful in industry, such as \emph{Photoshop}. The histogram of a hazy image typically peaks around a specific value, with few voxels close to zero, while the histogram of a clear image is more evenly distributed from 0 to 255. \cite{chi_single_2020} learns the ground truth histogram distribution with a full connection network, but it lacks spatial guidance. We aim to guide our network's attention on the channel dimension with a histogram by a one-dimensional vector.

\begin{figure*}[!ht]
    \centering
    \includegraphics[width=0.8\linewidth,trim= 0 0 0 0,clip]{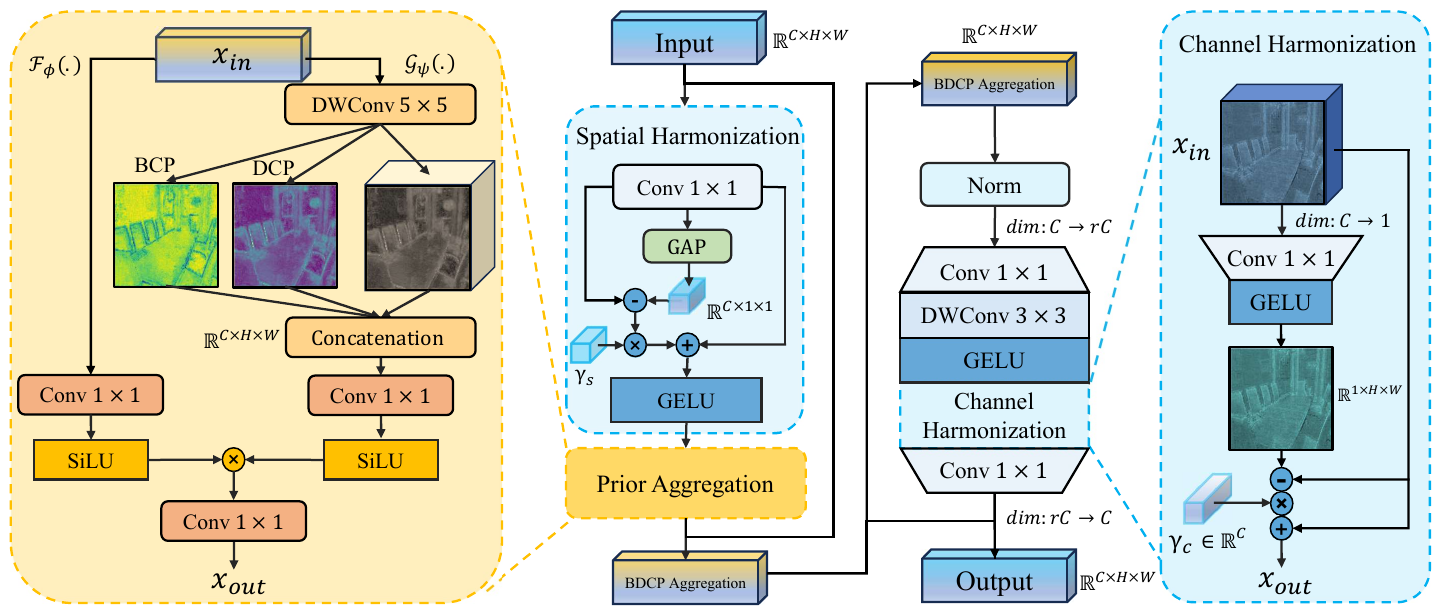}
    
    \caption{
    Structure of the encoder and decoder blocks:
    Spatial Harmonization Module $\mathrm{SH}(\cdot)$,  Prior Aggregation Module $\mathrm{PA}(\cdot)$, and  Channel Harmonization Module $\mathrm{CH}(\cdot)$ are cascaded. $\mathrm{SH}(\cdot)$ and $\mathrm{PA}(\cdot)$ combine to aggregate spatial information.
    }
    \label{fig:moga_multiorder}
\end{figure*}

\section{Proposed Method}
\label{sec:method}
In this section, we first present the overall architecture of PGH$^2$Net, shown in Fig.~\ref{fig:pipeline}. Following this, we describe the core components of PGH$^2$Net: spatial harmonization module, prior aggregation module, channel harmonization module, sandwich module, and histogram equation guide module (HEGM). Finally, the training loss function is defined.

\subsection{Overall Architecture}
As shown in Fig.~\ref{fig:pipeline}, the proposed PGH$^2$Net uses triple levels of architecture to efficiently learn hierarchical representations. Both the encoder and decoder networks comprise three scales. Specifically, given a hazy image of dimensions $3 \times H \times W$, a convolutional layer with a kernel size of $3\times3$ is applied to extract shallow features. These shallow features, which have dimensions $C \times H \times W$, then pass through a three-scale symmetric encoder-decoder structure. Transformation yields enhanced features with comprehensive image information through $n$ stages. Each stage includes prior aggregation, spatial harmonization, and channel harmonization, illustrated in Fig.~\ref{fig:moga_multiorder}. To integrate our proposed sandwich module, we mount it in the deepest layer since the BCP and DCP information is easier to learn in the shallow feature extractor but harder to learn as layers increase. Therefore, the HEGMs are utilized among different layers since the HE remains constant in different sizes.

\subsection{Spatial Aggregation Block}
\label{sec:method_sa}
We propose the spatial aggregation block (SA) to learn the Harmonized representations of B/DCP by a pure convolutional design, as shown in Fig.~\ref{fig:moga_multiorder} (left part), which consists of two cascaded components. 

\paragraph{Spatial Harmonization.}
According to our proposed harmonization prior, we extract diverse features with both \textit{static} and \textit{adaptive} locality perceptions in the SH module.
Since convolutions are inherently high-pass filters~\citep{iclr2022how, wang2022anti},
there are two complementary counterparts, fine-grained local texture and complex global shape, which are instantiated by $\mathrm{Conv}_{1\times 1}(\cdot)$ and $\mathrm{GAP}(\cdot)$, respectively. To counter the network's inherent interaction bias strengths~\citep{li2023moganet}, we design $\mathrm{SH}(\cdot)$ to adaptively exclude the trivial (overlooked) interactions, defined as:
\begin{align}
    \label{eq:FD_proj}
    \Ymat &= \mathrm{Conv}_{1\times 1}(\Xmat),\\
    \label{eq:FD}
    \Zmat &= \mathrm{GELU}\Big(\Ymat + \gammav_{s}\otimes\big(\Ymat - \mathrm{GAP}(\Ymat)\big) \Big),
\end{align}
where ${\gamma}_{s} \in \mathbb{R}^{C\times 1}$ denotes a scaling factor initialized as zeros. Reweighting the complementary interaction component $\Ymat - \mathrm{GAP}(\Ymat)$, $\mathrm{SH}(\cdot)$ also increases spatial feature diversities~\citep{iclr2022how, wang2022anti}.

\paragraph{Prior Aggregation.}
Since the difference between clean and degraded images of B/DCP in Fig.~\ref{fig:motivation_SM}, the associated priors and sparse constraints aid in restoring clear images. Then, we ensemble the B/DCP and local edge features in the context branch and adaptively select the informative channels by the gating aggregation in the PA module. The establishment of this prior principle is predicated upon empirical observation~\cite{he2010single, yan_image_2017, yao2024neural},  revealing that within the vast majority of patches in natural scenes, there consistently exists a feature tensor in which the highest and lowest intensity values of voxels tend to exhibit a pronounced prominence.  In this paper, the B/DCP of the feature domain is defined by 
\begin{align}
\label{eq:BD_prior}
\mathrm{D}(\Fmat)(\xv) & =\min _{\yv \in \Omega(\xv)}\left(\min _{c \in\{0...N\}} \Fmat ^c(\yv)\right), \\
\mathrm{B}(\Fmat)(\xv) & =\max _{\yv \in \Omega(\xv)}\left(\max _{c \in\{0...N\}} \Fmat ^c(\yv)\right),
\end{align}

where $\yv$ is the patch of location $\xv$ in the feature map, $\Fmat ^c$ is a channel of $\Fmat$, and $\Omega(\xv)$ denotes a local patch centered at $\xv$. $N$ denotes the number of channels.

Unlike previous work that simply combined DWConv with self-attention to model local and global interactions~\citep{eccv2022edgeformer, nips2022hilo, nips2022iformer, nips2022hornet,huang2023learned, guo2025mambair,yao2023towards,guan2023mutual}, we employ three different branches in parallel to capture DCP, BCP and identical interactions: Given the input feature $\Xmat \in \mathbb{R}^{C\times HW}$, $\mathrm{DW}_{5\times 5, d=1}$ is first applied for extracting features; then, the output is factorized into $\mathrm{B}({\Xmat})$, $\mathrm{D}(\Xmat)$ and an identical mapping $\Xmat$; finally, the outputs are concatenated to form B/DCP contexts, $\Ymat_{C} = \mathrm{Concat}(B({\Xmat}), D(\Xmat), \Xmat)$. 

After injecting B/DCP into feature maps, we utilize the gating aggregation to adaptively fuse priors and contextual features (\textit{e.g.,} edges). Taking the output from $\mathrm{SH}(\cdot)$ as the input, the output of the Prior Aggregation is written as:
\begin{align}
    \label{eq:moga}
    \Zmat &= \underbrace{\mathrm{SiLU}\big( \mathrm{Conv}_{1\times 1}(\Xmat) \big)}_{\mathcal{F}_{\phi}} \otimes \underbrace{\mathrm{SiLU}\big( \mathrm{Conv}_{1\times 1}(\Ymat_{C}) \big)}_{\mathcal{G}_{\psi}}.
\end{align}
It produces informative representations with similar parameters and FLOPs as $\mathrm{DW}_{7\times 7}$ in ConvNeXt, which is beyond the reach of existing methods.
\subsection{Channel Harmonization Block}
\label{sec:method_ch}
Due to channel redundancy~\citep{Woo_2018_ECCV, iccv2019GCNet, icml2019efficientnet, cvpr2020Orthogonal}, vanilla MLP needs numerous parameters ($r$ default to 4 or 8) for optimal performance. To overcome this, most methods insert a channel enhancement module, \textit{e.g.,} SE module~\citep{hu2018squeeze}. We propose a lightweight channel harmonization module $\mathrm{CH}(\cdot)$ to reallocate channel-wise features in high-dimensional spaces, further developing it into a channel harmonization (CH) block. 
As shown in Fig.~\ref{fig:moga_multiorder},
\begin{equation}
\begin{aligned}
    \Ymat &= \mathrm{GELU}\Big(\mathrm{DW_{3\times 3}}\big(\mathrm{Conv_{1\times 1}}(\mathrm{Norm}(\Xmat))\big)\Big),\\
    \Zmat &= \mathrm{Conv_{1 \times 1}}\big(\mathrm{CH}(\Ymat)\big) + \Xmat.
\end{aligned}
\end{equation}
Concretely, $\mathrm{CH}(\cdot)$ is implemented by a channel-reducing projection $W_{r}: \mathbb{R}^{C\times HW}\rightarrow \mathbb{R}^{1\times HW}$ and GELU to gather and reallocate channel-wise information:
\begin{equation}
    \mathrm{CH}(\Ymat) = \Ymat + \gammav_{c}\otimes\big(\Ymat - \mathrm{GELU}(\Ymat\Wmat_{r})\big),
\end{equation}
where $\gamma_{c}$ is the channel-wise scaling factor initialized as zeros. It harmonizes the channel-wise feature with the complementary interactions $(\Ymat - \mathrm{GELU}(\Ymat\Wmat_{r}))$. 

\subsection{Sandwich Module}
\label{sec:method_sandwich}
The higher sparsity of dark and bright channels in a sharp image compared to a degraded image, along with sparse constraints from these priors, aids in the restoration of clear images. The proposed sandwich module can regularize the spatial attention space. Specifically, there are two branches in the sandwich module, and the main branch squeezes feature map $\Fmat$ from the encoder along the channel dimension with B/DCP, shown in Fig.~\ref{fig:pipeline}. In addition, the average pooling $\mathrm{GAP}(\cdot)$ is used for complementary degradation locations. Subsequently, a concatenation operation is used to merge these three feature maps $\mathrm{B}(\Fmat)$, $\mathrm{D}(\Fmat)$ as Eq.~(\ref{eq:BD_prior}), and $\mathrm{GAP}(\Fmat)$, facilitating the integration of representations from the degraded image and the B/DCP as follows:
\begin{align}
\Fmat_{B}=\operatorname{Conv}([\mathrm{B}(\Fmat), \mathrm{GAP}(\Fmat)]), \\
\Fmat_{D}=\operatorname{Conv}([\mathrm{D}(\Fmat), \mathrm{GAP}(\Fmat)]), 
\end{align}
As each channel exhibits distinct degradation patterns, we proceed to create channel-specific representations by applying channel-separated transformations to the input feature $\Fmat$ using depth-wise convolutions, followed by modulation as:
\begin{equation}
\Fmat_s=\operatorname{DW}(\Fmat) \otimes \left( \Fmat_{B} + \Fmat_{D}\right)+\operatorname{DW}(\Fmat),
\end{equation}
To incorporate constraints based on dark and bright channel priors into a network, we also utilize a $\textit{l}_1$-regularization term to enforce sparsity during training.

\subsection{Histogram Equation Guided Module}

As another parallel branch to B/DCP in the bottleneck of the network, we introduce an innovative approach for single-image dehazing utilizing HEGM. Unlike other attention mechanism that operates on the 2D feature map, our model takes 1D histogram equation distribution data, specifically histograms in the image domain, as guided input. This phenomenon is visually demonstrated in Fig.~\ref{fig:motivation_HE}. Additionally, all guided features in the intermediate layers are 1D, simplifying our model and improving the ease of training compared to other CNN-based methods.

Take into account a feature tensor $\Fmat$, where $n_i$ represents the number of voxel value $i$. The probability of encountering a voxel with level $i$ in the image is the ratio of occurrences
\begin{equation}
    p_{\Fmat}(i)={\frac {n_{i}}{n}} \quad 0\leq i<L,
    \label{eq:prob}
\end{equation}
where $L$ is typically $256$ and $n$ is the total number of voxels. Cumulative distribution function (CDF) is defined as
\begin{equation}
    {cdf_{\Fmat}(i)=\sum _{j=0}^{i}p_{\Fmat}(j)},
\end{equation}
which is also the feature's accumulated normalized histogram. The general histogram equalization formula is:
\begin{equation}
    {\Hmat_{\Bar{\Fmat}}(i)=\mathrm {round} \left({\frac {{cdf}_{\Fmat} (i)-\ {cdf}_{\min }}{1 - {cdf} _{\min}}}\times (L-1)\right)}
\end{equation}
where $\Hmat_{\Bar{\Fmat}}(i)$ is the remapped voxel value in the new equalized image, $cdf_{\min}$ is the minimum non-zero value of the cumulative distribution function, and $L$ is the number of levels being used. Then we obtain a new image $\Bar{\Fmat}$ with histogram equalization. New probability $p_{\Bar{\Fmat}}$ is calculated by Eq.~\ref{eq:prob}.

As illustrated in Fig.~\ref{fig:pipeline}(c), the feature voxel $\Fmat$ is split along the channel, and the histogram of each channel is calculated and normalized to probability $p_{c}$ individually. Match feature voxel probability to $p_{\Bar{\Fmat}}$ so that the histogram of each channel generates a corresponding score with Cosine similarity: 
\begin{equation}
\Fmat ^ {\prime\prime}=\operatorname{Norm}(\operatorname{Sim}(p_{\Fmat},  p_{\Bar{\Fmat}}) ) \times \operatorname{Norm} \left(\operatorname{GAP}(\Fmat)\right),
\end{equation}
where $\operatorname{Sim}$ is Cosine similarity. After broadcasting, the score weights attention map is used to element-wise multiply with input $\Fmat$, and residual addition, which is formulated as: 
\begin{equation}
{\Fmat_H} = \Fmat \otimes \Fmat^{\prime\prime} + \Fmat.
\end{equation}

\begin{table*}[!t]
\renewcommand{\arraystretch}{1.0}
\setlength\tabcolsep{0.1mm}
\begin{center}
\begin{tabular}{l|c|cc|cc|cc|cc|cc}
\toprule
 & &\multicolumn{2}{c|}{SOTS-Indoor} & \multicolumn{2}{c|}{SOTS-Outdoor} & \multicolumn{2}{c|}{Dense-Haze} & 
 \multicolumn{2}{c|}{O-HAZE} & Params & FLOPs\\ 
Method  & Venue & PSNR ↑ & SSIM ↑ & PSNR ↑& SSIM ↑& PSNR ↑& SSIM ↑& PSNR ↑& SSIM ↑ & (M) & (G) \\ \midrule
DehazeNet~\cite{DehazeNet} & TIP'16 & 19.82 & 0.821 & 24.75 & 0.927 & 13.84 & 0.43 & 17.57 & 0.77 & 0.009 & \textbf{0.581} \\
AOD-Net~\cite{AODNet}  & ICCV'17 & 20.51 & 0.816 & 24.14 & 0.920 & 13.14 & 0.41 & 15.03 & 0.54 & \textbf{0.002} & 0.115 \\
GridDehaze~\cite{liu2019griddehazenet}  & CVPR'19 & 32.16 & 0.984 & 30.86 & 0.982 & - & -  & - & - & 0.956 & 21.49 \\
MSBDN~\cite{MSBDN} & CVPR'20 & 33.67 & 0.985 & 33.48 & 0.982 & 15.37 & 0.49 & 24.36 & 0.75 & 31.35 & 41.54 \\
FFA-Net~\cite{qin2020ffa} & AAAI'20 & 36.39 & 0.989 & 33.57 & 0.984 & 14.39 & 0.45 & 22.12 & 0.77 & 4.456 & 287.8 \\
AECR-Net~\cite{AECRNet} & CVPR'21 & 37.17 & 0.990 & -& - & 15.80 & 0.47 & - & - & 2.611 & 52.20 \\
DeHamer~\cite{dehamer} & CVPR'22& 36.63 & 0.988 & 35.18 & 0.986 & 16.62 & 0.56 & - & - & 132.45 & 48.93 \\
PMNet\cite{PMNet} &ECCV'22 & 38.41 & 0.990 & 34.74 & 0.985 & 16.79 & 0.51 & 24.64 & 0.83 & 18.90 & 81.13 \\ 
MAXIM-2S~\cite{MAXIM}& CVPR'22& 38.11& 0.991& 34.19& 0.985& -& -& -& -& 14.10&216.0\\
DehazeFormer\cite{song2023vision}& TIP'23& 38.46& 0.994& 34.29& 0.983& 16.29& 0.51& -& -& 14.10&216.0\\
TaylorFormer~\cite{qiu_mb-taylorformer_2023}& ICCV'23&  40.71&  0.992&  37.42 &  0.989&  16.66&  0.56&   25.05& 0.788& 2.68& 38.50\\
LH-Net~\cite{yuan_lhnet_2023}& MM'23& 37.04& 0.989& 36.05& 0.986& 18.87 & 0.561 & -& -& 35.64& - \\
MITNet~\cite{shen_mutual_2023} & MM'23 & 40.23 & 0.992 & 35.18 & 0.988& 16.97 & 0.606 & -&- &2.83 & 16.25 \\
 DEA~\cite{DEA} & TIP'24& 41.31& 0.995& 36.59& 0.989& -& -& -& -& 3.65&32.23 \\ 
\midrule
\rowcolor[rgb]{0.898,0.898,0.902}
\textbf{PGH$^2$Net} & \textbf{Ours} & \textbf{41.70} & \textbf{0.996} & \textbf{37.52} & \textbf{0.989} & \textbf{17.02}& \textbf{0.61} & \textbf{25.47}& \textbf{0.88} & 1.76 & 16.05 \\ \bottomrule
\end{tabular}

\end{center}
\caption{Image dehazing results on both synthetic dataset~\cite{RESIDE} and real-world datasets~\cite{dense-haze,Ohaze}.
} 
\label{tab:dehazing}
\end{table*}
%
\subsection{Triple Learning Objective}
To facilitate the sparsity of B/DCP, we adopt $\ell_1$ loss in the spatial, frequency, and structural domains:
\begin{align}
&\mathcal{L}_{\text {spatial }}=\sum_{i=1}^3 \frac{1}{P_i}\left\|\hat{\mathbf{X}}_i-\mathbf{X}_i\right\|_1,\\
&\mathcal{L}_{\text {frequency }}=\sum_{i=1}^3 \frac{1}{P_i}\left\|\mathrm{F}\left(\hat{\mathbf{X}}_i\right)-\mathrm{F}\left(\mathbf{X}_i\right)\right\|_1, \\
&\mathcal{L}_{\text {ssim}}=\sum_{i=1}^3 \frac{1}{P_i}\left\| \mathrm{SSIM}\left(\hat{\mathbf{X}}_i\right)-\mathrm{SSIM}\left(\mathbf{X}_i\right)\right\|_1, 
\end{align}
where $i$ is the index of multiple outputs, as illustrated in Fig.~\ref{fig:pipeline}(a); \(\hat{\mathbf{X}}\) and \(\mathbf{X}\) denote the predicted image and ground truth, respectively; $P_i$ represents the total elements of the image for normalization, and $\mathrm{F}$ represents the fast Fourier transform. 
Structural Similarity Index Measure (SSIM) to measure the local structural similarity between images or patches, which is formulated as the combination of three metrics: luminance, contrast, and structure. 
The final loss function is given by the three terms:
\begin{equation}
    \mathcal{L}_{\text {total}}=\mathcal{L}_{\text {spatial}}+\lambda_1 \mathcal{L}_{\text {frequency}} + \lambda_2 \mathcal{L}_{\text {ssim}},
\end{equation}
where loss weight $\lambda$ requires fine-tuning in practice and we used $\lambda_1=0.5$ and $\lambda_2=1$ as the final setting.

\section{Experimental Results}
In this section, we evaluate our proposed PGH$^2$Net in four data sets for image dehazing tasks, including indoor synthetic data, outdoor synthetic data, and two real data. 

\subsection{Datasets and Evaluation Metrics}

\noindent\textbf{Evaluation.} We calculate the Peak Signal-to-Noise Ratio (PSNR), and Structural Similarity index (SSIM)~\cite{wang2004image} between the predicted results and ground-truth images for all datasets. Floating point operations (FLOPs) are measured on the patch of size $3\times 256\times 256$. 

\begin{figure}[!t]
    \centering
    \includegraphics[width=\linewidth]{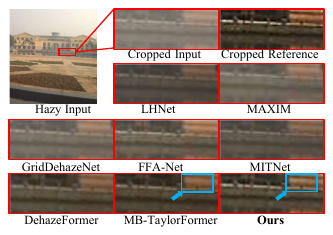}
    \caption{Image dehazing comparisons on the SOTS-Outdoor~\cite{RESIDE} test sets. The \textcolor{red}{red} box is zoomed in by 6\texttimes~for visualization. The \textcolor{cyan}{cyan} arrows point to the superior performance of our method.}
    \label{fig:sots}
\end{figure}

\begin{figure}[!t]
    \centering
    \includegraphics[width=\linewidth]{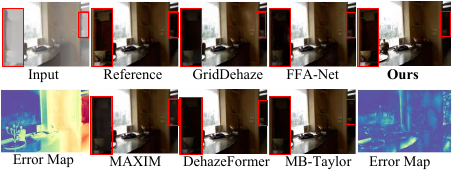}
    \caption{Comparisons on the Indoors~\cite{RESIDE} data.}
    \label{fig:its}
\end{figure}

\begin{figure*}[!htbp]
    \centering
    \includegraphics[width=0.95\linewidth]{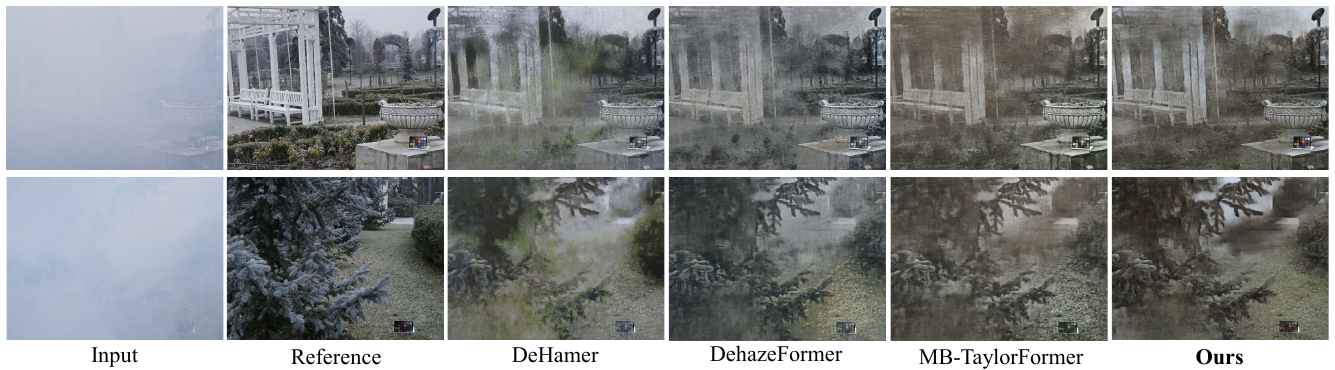}
    \caption{Comparisons on the Dense-Haze~\cite{dense-haze}.}
    \label{fig:real}
\end{figure*}

\noindent\textbf{Image Dehazing Datasets.} We train and evaluate our models on synthetic and real-world datasets for image dehazing. Following~\cite{wang2022cycle, wang2024ucl}, we train separate models on the RESIDE-Indoor and RESIDE-Outdoor datasets~\cite{RESIDE}, and evaluate the resulting models on the corresponding test sets of RESIDE, \textit{i.e.,} SOTS-Indoor and SOTS-Outdoor, respectively. In addition, we adopt two real-world datasets, \textit{i.e.,} Dense-Haze~\cite{dense-haze} and O-HAZE~\cite{Ohaze}, to verify the robustness of our model in more challenging real-world scenarios. 
\subsection{Implementation Details} 

The models are trained using Adam~\cite{kingma2014adam} with initial learning rate as $8e^{-4}$, which is gradually reduced to $1e^{-6}$ with cosine annealing~\cite{loshchilov2016sgdr}. For data augmentation, we adopt random horizontal flips with a probability of 0.5. Models are trained on 32 samples of size $256\times 256$ for each iteration.

\subsection{Image Dehazing Results}
\textbf{Quantitative Comparisons.} We report the quantitative performance of image dehazing approaches on both synthetic~\cite{RESIDE} and real-world~\cite{dense-haze, Ohaze} datasets in Tab.~\ref{tab:dehazing}. Overall, our method receives higher performance on all datasets than other state-of-the-art algorithms. Specifically, on the daytime synthetic dataset SOTS-Indoor~\cite{RESIDE}, our method outperforms MB-TaylorFormer~\cite{qiu_mb-taylorformer_2023} by 0.99 dB PSNR with only 42\% parameters and 67\% FLOPs. Furthermore, our model yields a significant performance gain of 3.23 dB in terms of PSNR over Transformer model DeHamer~\cite{dehamer} on SOTS-Outdoor~\cite{RESIDE} with fewer parameters. 

\noindent\textbf{Visual Comparisons in synthetic dataset.} The daytime visual results produced by several dehazing methods are illustrated in Fig.~\ref{fig:sots},~\ref{fig:its}. Our method is more effective in removing haze blurs in both indoor and outdoor scenes than other algorithms, such as blurs on the doors in the top two images of Fig.~\ref{fig:sots}.
The proposed PGH$^2$Net can retrieve not only the sharp shapes of the objects but also the colorful, fine textures and details. 
Simultaneously, PGH$^2$Net avoids noise and artefacts in the background that appear in other methods.

\noindent\textbf{Visual Comparisons in a real-world dataset.}
In Fig.~\ref{fig:real}, our method is well generalized to the more challenging real-world scenarios following USCFormer~\cite{wang2023uscformer} and obtains the best performance. Our method recovers details of the grasses and trees. The color plate at the lower right corner indicates color correction ability of our method.

\subsection{Ablation Studies}
We conduct ablation studies to demonstrate the effectiveness of our modules by training the model on RESIDE-Indoor~\cite{RESIDE} and testing on SOTS-Indoor~\cite{RESIDE}. The model is trained with an initial learning rate of $8e^{-4}$ and a batch size of $32$, ending in epoch $100$. 


\noindent\textbf{Effects of the encoder and decoder module.}
We first ablate the spatial aggregation module and the channel aggregation module CH$(\cdot)$ in Tab.~\ref{tab:ablation}. Spatial modules include SH\textbf{$(\cdot)$} and PA\textbf{$(\cdot)$}, containing the gating branch. We found that all proposed modules yield improvements with favorable costs.

\begin{table}[htbp]
    \setlength{\tabcolsep}{0.5mm}
    \centering
        \begin{tabular}{l|clcc}
        \toprule
        Modules & PSNR↑ &SSIM↑ & Params. (M) & FLOPs (G) \\ \hline
        ~~Baseline & 35.64  &0.985& 1.34 & 15.96 \\
        +Gating branch & 36.27  &0.987& 1.68 & 15.98 \\
        +$\mathrm{PA(\cdot)}$ & 37.14  &0.987& 1.72& 16.00 \\
        +$\mathrm{SH(\cdot)}$ & 37.48  &0.990& 1.75& 16.01 \\
        \rowcolor[rgb]{0.898,0.898,0.902} 
        +$\mathrm{CH(\cdot)}$ & \bf{38.00}  &\bf{0.992}& 1.76 & 16.05 \\ \hline
        \end{tabular}
    \caption{Ablation of our modules on the SOTS-Indoor~\cite{RESIDE} dataset. The baseline uses the non-linear projection and $\mathrm{DW}_{5\times 5}$ as $\mathrm{SH}(\cdot)$ and the  MLP as $\mathrm{CH}(\cdot)$.
    }
    \label{tab:ablation}
\end{table}

\begin{table}[t!]
\begin{center}
\setlength\tabcolsep{0.1mm}
\begin{tabular}{@{}ccccccc@{}}
\toprule
 & Sandwich& HEGM & PSNR↑ & SSIM↑ & Params. (M) & FLOPs (G) \\ 
 \midrule
(a) &  &  & 34.69  &0.985& 1.74 & 15.98 \\
(b) & \checkmark &  & 36.09  &0.989& 1.76 & 16.04 \\
(c) &  & \checkmark & 35.32  &0.985& 1.74 & 16.00 \\
\rowcolor[rgb]{0.898,0.898,0.902} 
(d) & \checkmark & \checkmark & \textbf{38.00}  &\bf{0.992}& 1.76 & 16.05 \\
\toprule
\end{tabular}
\end{center}
\caption{Ablation studies for different bottleneck attention of PGH$^2$Net on the SOTS-Indoor~\cite{RESIDE}. }
\label{tab:breakdown}
\end{table}

\noindent\textbf{Effects of bottleneck attention module.}
As shown in Tab.~\ref{tab:breakdown}a, the baseline receives 34.69 dB PSNR. sandwich module (Tab.~\ref{tab:breakdown}b) and HEGM (Tab.~\ref{tab:breakdown}c) yield accuracy gains of 1.40 and 0.63 dB over the baseline, respectively. 



In addition, the visual results of our sandwich module $\mathrm{SM}(\cdot)$ are illustrated in Fig.~\ref{fig:motivation_SM}. The sandwich module helps the model focus more on the severe degradation regions, \textit{e.g.,} metal fence. HEGM further highlights the accurate voxel value distribution (see Fig.~\ref{fig:motivation_HE}). 


\section{Conclusion}
In this paper, we present a triple priors guided network for image dehazing, dubbed PGH$^2$Net, which is effective and computationally efficient. To our knowledge, this is the first work to reveal the relationship between B/DCP and spatial guidance in hierarchical feature representation. We are the first to utilize HE matching similarity to harmonize the channel-wise features, which is effective and low-cost. By collaborating with triple priors, PGH$^2$Net can leverage their individual strengths and provide complementary information in a harmonious manner, shown in Fig.~\ref{fig:motivation_SM}(d-f).
Our future work will explore the framework in other image tasks.
\newpage
\section{Acknowledgements}
This work was supported by the National Key R$\&$D Program of China (grant number 2024YFF0505603, 2024YFF0505600), the National Natural Science Foundation of China (grant number 62271414), Zhejiang Provincial Outstanding Youth Science Foundation (grant number LR23F010001), Zhejiang “Pioneer" and “Leading Goose"R$\&$D Program(grant number 2024SDXHDX0006, 2024C03182), the Key Project of Westlake Institute for Optoelectronics (grant number 2023GD007), the 2023 International  Sci-tech Cooperation Projects under the purview of the “Innovation Yongjiang 2035” Key R$\&$D Program (grant number 2024Z126) and the Zhejiang Province Postdoctoral Research Excellence Funding Program (grant number ZJ2024086).
\appendix

\bibliography{aaai25}

\end{document}